\documentclass{article}

\usepackage{PRIMEarxiv}

\usepackage[utf8]{inputenc} 
\usepackage[T1]{fontenc}    
\usepackage{url}            
\usepackage{booktabs}       
\usepackage{amsfonts}       
\usepackage{nicefrac}       
\usepackage{microtype}      
\usepackage{lipsum}
\usepackage{fancyhdr}       
\usepackage{graphicx}       
\graphicspath{{media/}}     
\usepackage[affil-it]{authblk}
\usepackage{float}
\usepackage{chngcntr}

\usepackage{hyperref}       

\title{A Framework for Monitoring and Retraining Language Models in Real-World Applications
}

\author{Jaykumar Kasundra}
\author{Claudia Schulz}
\author{Melicaalsadat Mirsafian}
\author{Stavroula Skylaki}
\affil{Thomson Reuters Labs}

\begin{document}
\maketitle

\begin{abstract}
In the Machine Learning (ML) model development lifecycle, training candidate models using an offline holdout dataset and identifying the best model for the given task is only the first step. After the deployment of the selected model, continuous model monitoring and model retraining is required in many real-world applications. There are multiple reasons for retraining, including data or concept drift, which may be reflected on the model performance as monitored by an appropriate metric. Another motivation for retraining is the acquisition of increasing amounts of data over time, which may be used to retrain and improve the model performance even in the absence of drifts. We examine the impact of various retraining decision points on crucial factors, such as model performance and resource utilization, in the context of Multilabel Classification models. We explain our key decision points and propose a reference framework for designing an effective model retraining strategy.
\end{abstract}


\section{Introduction}

The typical model development lifecycle consists of four phases: 1) problem scoping, 2) data definition and collection, 3) model training and iterative improvement through error analysis, and 4) model deployment in production and implementation of continuous monitoring and retraining 
\cite{nigenda_amazon_2022}. While the first three phases are typically performed in an offline setting, model deployment represents the critical step where the ML model becomes available in a production environment, a live application, where it needs to process live data and ideally sustain performance over time to keep delivering value.

Model monitoring refers to the process of evaluating the quality of the production data and the performance of the model according to relevant metrics over time. When either data quality or model performance does not meet predefined criteria, a monitoring warning can be triggered, to alert the model owners. Defining an effective model monitoring and retraining strategy is key to successful ML model deployment since it can safeguard model quality over prolonged periods of time. Model monitoring can detect model drift, a degradation of the model’s prediction quality due to changes in the environment  \cite{bayram_concept_2022}. Model drift manifests in various ways, such as changes in the properties of the dependent variables (concept drift), changes in the properties of the independent variables (data drift), or changes in the upstream data pipeline \cite{ackerman_automatically_2021, yu_automatic_2021}. Model retraining can be used to remedy degrading model performance whenever detected by model monitoring. In addition, as data accumulates over time, retraining models with newly acquired data may improve model quality. In this scenario, retraining is a proactive approach towards deploying improved models even in the absence of monitoring warnings. 

\begin{figure}[htp]
    \centering
    \includegraphics[width=7.5cm]{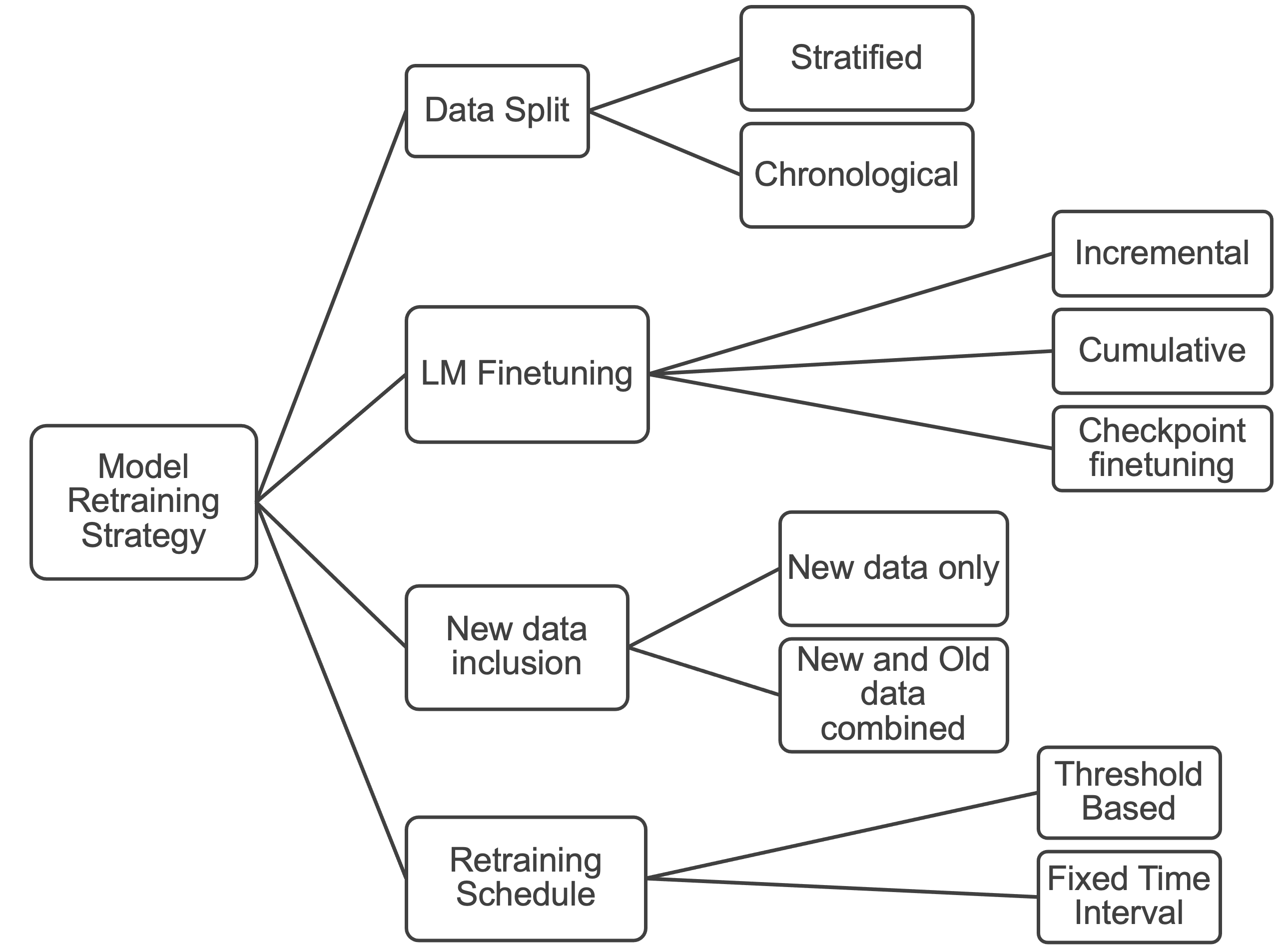}
    \caption{Model Retraining Decision Points}
    \label{fig:Model_Retraining_Decision_Points}
\end{figure}

The idea to automatically and continuously retrain a model to adapt to changes that might occur in the data is called ‘Continuous Training’ (CT). 
There are different approaches to perform CT, each with its own pros, cons, and cost. Some important questions  to address when designing a CT strategy for the deployed model – hereinafter referred to as the champion - relate to 1) the way the data are split (e.g., stratified split, which aims to ensure that the distribution of labels and their relations in the data is balanced across different data splits, vs. chronological split) and how new data is incorporated during retraining, 2) the way the Language Model (LM) is finetuned  (e.g., from scratch or from the champion), and finally 3) the retraining schedule and the retraining trigger (to ensure optimal quality and relevance, while also considering cost efficiency) (see Figure~\ref{fig:Model_Retraining_Decision_Points}).

In this study, we shed light on the above questions by experimenting with different retraining setups for Multilabel Classification (MLC). MLC is very common NLP use case and often comes with challenges \cite{katz2023natural}. For example, the class frequencies tend to be very imbalanced. A large number of classes often leads to many classes with very little data and a few classes with a lot of data
\cite{rubin2012statistical}. An additional challenge is class fluctuation over time, in other words, temporal concept drift \cite{chalkidis-sogaard-2022-improved}.

The use of large pre-trained LMs for MLC is very common because of the significantly better performance of these models on the MLC benchmarks \cite{gonzalez-carvajal_comparing_2021}. Hence, we focus on the use case where MLC is tackled using LMs.

We define a set of key decision points that should be considered during the design of the monitoring and retraining strategy and provide empirical evidence on how these decisions affect the model’s prediction quality. Optimal decisions may save costs on retraining and help sustain model performance over time.

The contributions of our work are as follows:
\begin{enumerate}
\item A framework of decision points to develop a model retraining process.
\item Strategies for including newly acquired data for model retraining and evaluation.
\item Empirical evidence that frequent model finetuning can be an effective solution to address model performance drops and optimal resource utilization in data drift scenarios.
\end{enumerate}

\section{Related work}
Despite being a critical part of the model development lifecycle, optimal retraining strategies may be challenging to define. The retraining pipeline is often manually or automatically triggered by a monitoring alarm. There is a wealth of studies on how model monitoring delivers insights in different types of essential model supervision, e.g., outlier detection \cite{chandola_outlier_2009}, drift detection \cite{lu_learning_2019}, bias drift \cite{das_fairness_2021}, and adversarial attacks \cite{chakraborty_adversarial_2018}. In its most rudimentary form, monitoring can be a simple evaluation of the model performance within a time window, where if the model performance metric falls below a certain threshold, then a monitoring alarm is triggered.

While model monitoring may be used to answer the question “when should we retrain?”, the subsequent question would be “which data should we use to retrain?”. The data split strategy is an important decision for a CT setup. The retraining strategy should be able to effectively distribute data into train, validation, and test set. \cite{raykar_data_2015} proposed workflows to manage the allocation of newly acquired data into different sets in continuous model building and updating scenarios. The authors propose three different workflows (parallel dump, serial waterfall, and hybrid) for allocating new data into the existing training, validation, and test splits. Emphasis is laid on avoiding the bias due to the repeated use of the existing validation or test set.

Finally, how should we best continuously retrain LMs? There is a significant advance in NLP with the emergence of Transformer-based LMs. These LMs are utilized for downstream tasks with little to no “finetuning” on custom datasets \cite{ben-zaken-etal-2022-bitfit}. It is important, therefore, to understand the effect of retraining strategy for LMs and its impact on model performance.

\cite{schulz_analysis_2019} performed simulation to train models with only a small number of ‘already annotated’ texts and then continuously adjusting the models when ‘newly annotated’ texts become available. The experiments in the research involved BiLSTM network with a conditional random field output layer \cite{reimers_reporting_2017}. The authors showed that model performance for incremental training (training previous model with only the new data points) with sufficiently large number of new data points, cumulative training (training the previous model on new and old data combined), and retraining (training the model from scratch with old and new data points combined) have insignificant differences. 

The objective of the present work is to map out the key decision points when designing monitoring and retraining strategies for optimal performance of a LM, while providing empirical evidence on how these decisions can affect model performance over time in a real-world application. 

\section{Experiments}
\subsection{Data}

In this study, we simulate a realistic scenario of an end-to-end model development lifecycle and retraining strategy, in particular CT of an MLC model as it would be required if this model was deployed. For this reason, we select a timestamped dataset that has been used for document classification in a human-in-the-loop system. The dataset consists of different types of legal documents, such as policies, best practice recommendations, contracts, legal insights and news, etc., created by an in-house editorial team during the span of multiple years. We simulate the following project stages: 1) Research, which includes model development and evaluation, and 2) Model in deployment, which includes model monitoring and retraining. 

For the given dataset, our data range is Jan 1990 to Dec 2020. We simulate a research stage that considers data from Jan 1990 to Dec 2018 to train a model. 
The  model deployment, and therefore acquisition of user feedback and new data, is simulated using data from Jan 2019 to Dec 2020. The dataset comprises  234,877 documents, out of which 189,527 documents (documents received until Dec 2018) are utilized for training the model, while the remaining documents are used during the model deployment phase. The training dataset is split into train, validation, and test sets in a 70:15:15 ratio. The total number of target labels in the dataset is 35, and approximately 90\% of the documents have three or fewer labels. The average document length for the dataset is 1,080 words, with 65\% of documents containing less than 500 words.

\subsection{Methodology}
Model retraining strategies for the experiments are decided based on following important decision points.
\begin{enumerate}
    \item Data Splits strategy   for model training
        \begin{itemize}
            \item Stratified split - The target variable is used for iterative splitting to create the train, validation, and test datasets in a 70:15:15 ratio.
            \item chronological split - The data is split chronologically into training, validation, and test datasets in a ratio of 70:15:15, such that the oldest data is included in the training set, and the latest data is included in the test set.  
        \end{itemize}
    \item Strategy for LM Finetuning
        \begin{itemize}
            \item Incrementally finetune the champion model with mainly newly acquired data (some old data may be added)
            \item Cumulatively finetune the champion model with all previous data combined with newly acquired data
            \item Retrain the checkpoint model (LM Checkpoint like RoBERTa \cite{liu2019roberta}) with all previous data combined with newly acquired data
        \end{itemize}
    \item New data inclusion for model retraining
        \begin{itemize}
            \item Include just the newly acquired data
            \item Include newly acquired data as well as some or all of previous data
        \end{itemize}
    \item Retraining schedule
        \begin{itemize}
            \item Threshold based - Perform retraining when the model performance does not meet the minimum  model performance  criteria for a certain duration
            \item Fixed Time Interval - Perform retraining at a fixed interval
        \end{itemize}
\end{enumerate}

Decision points 1 and 3 both deal with training data, so some thought has to be put into the various ways in which one can use the previous data, newly acquired data and their combination to prepare the train, validation, and test dataset \cite{raykar_data_2015}.
Depending on how the initial data split was performed to build the initial model, there are 4 ways in which we prepare the retraining data.
\begin{enumerate}
    \item Stratified split \footnote{As implemented in the scikit-multilearn package \cite{2017arXiv170201460S}} with New data only (see Figure~\ref{fig:Stratified_split_with_rapid_data_refresh}): This strategy considers only the data that was acquired since the last model training/retraining and splits it into train, validation, and test datasets. This approach can be used to retrain the model on a small amount of newly acquired data.

    \item Stratified split with New and some old data combined (see Figure~\ref{fig:Stratified_split_with_waterfall_movement}): This strategy considers the data since last model training/retraining and splits the data into train, validation, and test dataset. It adds the entire validation dataset from previous model training/retraining into the current training dataset. It also adds x\footnote{Selecting the proportion is a design choice. Selecting higher value would make the model adapt to the recent data trends faster and vice versa. We implement the experiments with 0.5.} proportion of previous test dataset into current validation dataset and 1-x proportion of previous test dataset into current test dataset. This strategy is suitable for situations where the data undergoes gradual changes over time, and it is crucial for the model to avoid missing out on any data points. It is particularly effective when the goal is to develop a model with strong generalization capabilities, rather than one that is overly sensitive to specific temporary changes in the data.

    \item Chronological split with New data only (see Figure~\ref{fig:Chronological_split_with_rapid_data_refresh}): This strategy works similar to the Stratified split with New data only, except it splits the new data chronologically instead of stratified split. It is useful when one wants to evaluate the model’s performance on new data from different time periods.

    \item Chronological split with New and some old data combined (see Figure~\ref{fig:Chronological_split_with_moving_window}): In this strategy, the newly acquired data is split into train, validation, and test datasets chronologically, and the validation and test data from the previous model training/retraining are added to the new train dataset. This approach is advantageous when both the newly acquired and previous data need to be considered for model retraining.
    
\end{enumerate}

\subsection{Experimental Setup}
All experiments are performed using the RoBERTa pre-trained model \cite{liu2019roberta}. Detailed information about the finetuning process and associated hyperparameters can be found in Appendix~\ref{sec:appendix}.

For the Threshold based based retraining schedule, we set a threshold of 5\% relative drop in the model performance (weighted-f1 score), e.g. if the model performance on test data is 80\% then the threshold is set as 76\%. The model's performance is monitored by evaluating on the last 4 weeks of data every week. Retraining of the model is initiated when the threshold is breached consecutively for four times.
The retraining schedule based on a fixed time interval is set to 6 months since we received new data equivalent to 7-8\% of the initial training data size during a 6-month period.

We implement 12 retraining scenarios based on the decision points described above (see Figure~\ref{fig:Model_Retraining_Decision_Points}).
Each experiment consists of the following steps:
\begin{enumerate}
    \item Split the data into train, validation, and test datasets based on the split strategy during the research phase
    \item Train the champion model
    \item Monitor the model performance on monitoring dataset during the simulated deployment phase
    \item Identify the model retraining trigger date based on the model monitoring strategy (threshold based, or fixed interval based)
    \item Prepare the train, validation, and test datasets for model retraining based on the new data inclusion strategy
    \item Perform model retraining based on the LM finetuning strategy using the above datasets
    \item Compare the champion (currently deployed) and challenger (newly retrained) model performances on the new test dataset
    \item Continue step 3 to 7 until the end of the monitoring period
\end{enumerate}

\begin{table*}[htbp]
\centering
\resizebox{\linewidth}{!}{\begin{tabular}{|c|l|l|l|l|l|c|c|}
\hline
No. & Data Split	& LM Finetuning &	New data inclusion &	Retraining Schedule	 & \multicolumn{1}{|p{1.5cm}|}{\centering Retraining time} &	\multicolumn{1}{|p{2cm}|}{\centering No. of Retraining}	& \multicolumn{1}{|p{3cm}|}{\centering Avg. Performance on Monitoring Period} \\
\hline

\textbf{1} & \textbf{Stratified} & \textbf{Incremental} & \textbf{New data only} & \textbf{Threshold based} & \textbf{258 Mins} & \textbf{11} & \textbf{76.57} \\
2 & Stratified & Incremental & New data only & Fixed Interval & 106 Mins & 3 & 74.4 \\
3 & Stratified & Incremental & New + some old data & Threshold Based & 672 Mins & 9 & 75.35 \\
4 & Stratified & Incremental & New + some old data & Fixed Interval & 291 Mins & 3 & 72.43 \\
5 & Stratified & Cumulative & New + all old data & Fixed Interval & 1665 Mins & 3 & 69.7 \\
6 & Stratified & Checkpoint Finetuning & New + all old data & Fixed Interval & 3795 Mins & 3 & 74.58 \\
7 & Chronological & Incremental & New data only & Threshold Based & 53 Mins & 3 & 71.43 \\
8 & Chronological & Incremental & New data only & Fixed Interval & 101 Mins & 3 & 70.39 \\
9 & Chronological & Incremental & New + some old data & Threshold based & 248 Mins & 3 & 72.76 \\
10 & Chronological & Incremental & New + some old data & Fixed Interval & 144 Mins & 3 & 69.6 \\
11 & Chronological & Cumulative & New + all old data & Fixed Interval & 1908 Mins & 3 & 57.16 \\
12 & Chronological & Checkpoint Finetuning & New + all old data & Fixed Interval & 2894 Mins & 3 & 56.92 \\
\hline
\end{tabular}}
\caption{Model Retraining Strategy Results}
\label{tab:Model Retraining Strategy Results}
\end{table*}

\subsection{Results}

We are evaluating the efficacy of different model retraining techniques based on their average performance during the monitoring period. The results, as shown in Table \ref{tab:Model Retraining Strategy Results}, include the average performance during the monitoring period, the number of retraining cycles, and the time required for each retraining cycle. Our findings indicate that the most favorable results were obtained by training a model using a stratified split on the training data and performing regular finetuning with new data throughout the monitoring period. This suggests that Language Models (LMs) are capable of effectively learning and adapting to changes in the data when finetuned with a sufficient number of data points.

\section{Discussion}
\subsection{Data Split}
We observe that overall the performance of models using the stratified data split strategy is higher than when using the chronological strategy (see Table~\ref{tab:Model Retraining Strategy Results}). This can be attributed to the fact that the latter misses out on the most recent data during training because the most recent data is held-out as the test set. 

\begin{figure*}[ht]
    \centering
    \includegraphics[width=0.85\textwidth, height=4.5cm]{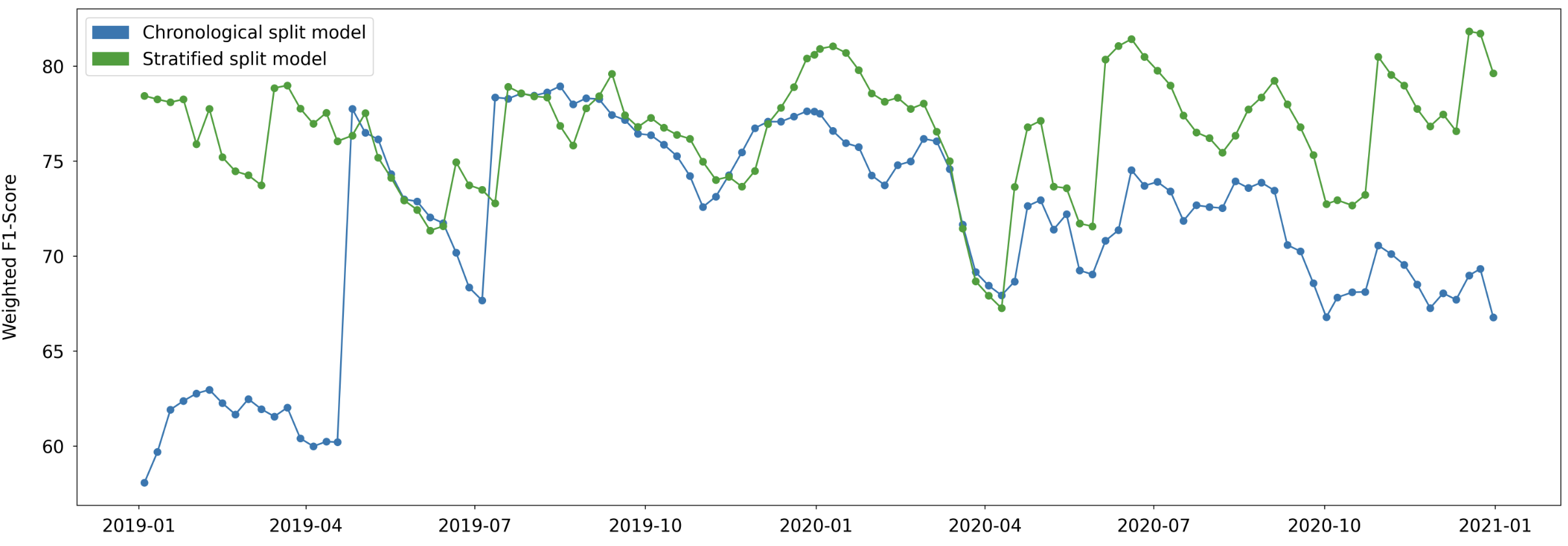}
    \caption{Performance of models developed with stratified vs. chronological split during monitoring period. Stratified split and chronological split refers to the strategies 1 and 7 from Table \ref{tab:Model Retraining Strategy Results} respectively}
    \label{fig:Stratified vs. Chronological Split}
\end{figure*}

To understand the effect of the data split strategy on model performance isolated from mixing new and old data, we compare models trained using stratified vs chronological splits, where retraining is done with only newly acquired data (so the original data is not re-used), represented by
strategies 1 and 7 in Table~\ref{tab:Model Retraining Strategy Results}.
Remember that the new data used for retraining is then split according to the same strategy as used during initial model training.
Table~\ref{tab:Model Retraining Strategy Results} and Figure~\ref{fig:Stratified vs. Chronological Split} illustrate that
training using a stratified split throughout considerably outperforms a chronological split.

\begin{figure*}[ht]
    \centering
    \includegraphics[width=0.85\textwidth, height=4.5cm]{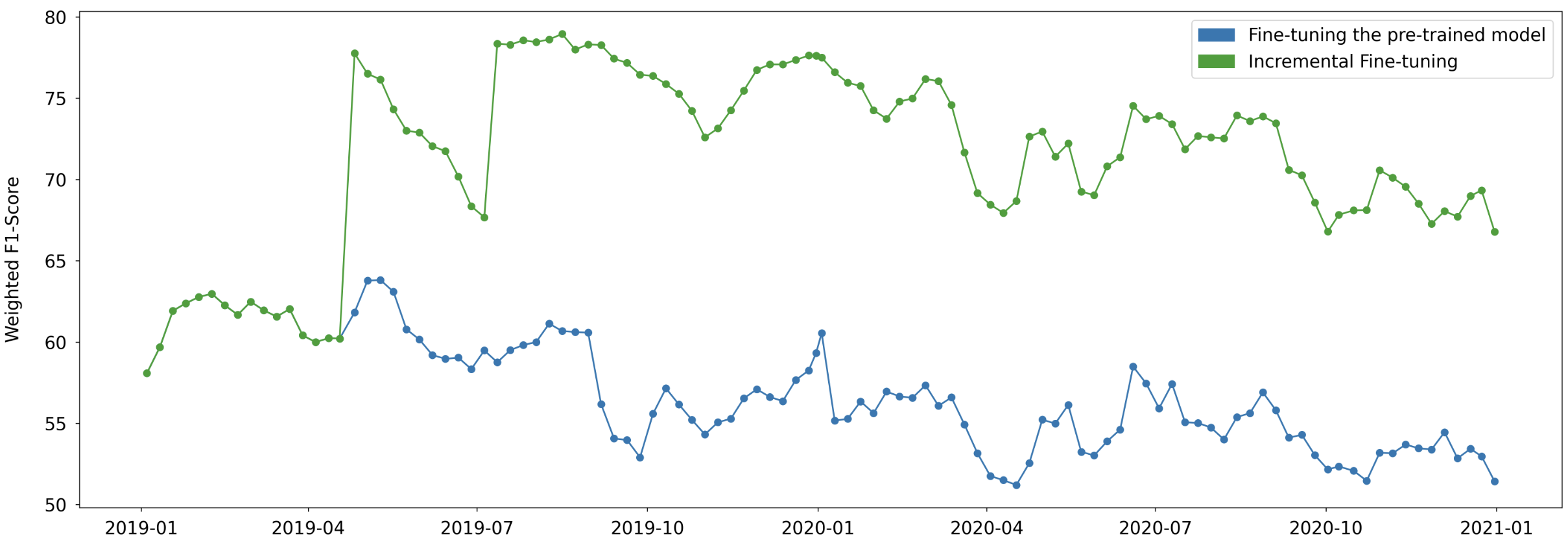}
    \caption{In the 'incremental finetuning' approach (Strategy 7 in Table \ref{tab:Model Retraining Strategy Results}), we finetuned the champion model on newly acquired data; in the 'finetuning the pre-trained model' approach (Strategy 12 in Table \ref{tab:Model Retraining Strategy Results}), we combined the newly acquired data with the old data and performed finetuning on the pre-trained RoBERTa base model.}
    \label{fig:Finetuning vs. Retraining from Scratch}
\end{figure*}

\subsection{LM Finetuning Strategy}
A common strategy for retraining is to finetune a pre-trained model with old data combined with newly acquired data and comparing with the current champion model. We observed that there is usually not much benefit in doing training/finetuning from scratch (Checkpoint Finetuning in Table~\ref{tab:Model Retraining Strategy Results}) in terms of model performance, while incurring much higher training time and therefore cost compared to incremental finetuning of the champion model. A similar pattern was observed for cumulative finetuning, which could not bring similar levels of model performance when compared to the Incremental approach. 

It is observed that it is better to incrementally finetune the champion model rather than going for finetuning the pre-trained model with all data for every retraining flag (Table \ref{tab:Model Retraining Strategy Results}). Incremental finetuning gave better performance and required lower training costs. Although catastrophic forgetting \cite{klabjan_neural_2020} can pose a problem in some cases, we found that incremental finetuning was not an issue in our case, as we observed data drift in our data and required our models to adapt to new data quickly.

The contrast between the two strategies was very prominent when the model was trained/retrained using a chronological split as we can see in Figure~\ref{fig:Finetuning vs. Retraining from Scratch}. We had observed temporal bias in our training dataset from research stage. We believe it to be the primary reason for the difference, and the retraining was insufficient to address this bias within the two-year monitoring period when the data was split chronologically. If the training dataset from the research stage did not have substantial bias, such a significant disparity in performance might not be observed.

\subsection{New Data Inclusion}
'New data only' strategy includes only the newly acquired data for model retraining and to create a test set for the champion vs. challenger model comparison. 

While 'New and some old data combined' strategy includes newly acquired data combined with some of the previous data to retrain model and to create a test set for performing champion vs. challenger model comparison. For stratified split, the 'New and some old data combined' strategy includes data from all previous time periods. This historical data representation dilutes over time and gets replaced with newer and fresh data. While for chronological split, the 'New and some old data combined' strategy is moving data window where the model will see validation and test data from previous period during the current period as part of training data. 

We observed that for stratified split, the 'New data only' strategy outperformed the 'New and some old data combined' strategy (Strategies 1 and 3 in Table \ref{tab:Model Retraining Strategy Results}). The data during the monitoring period exhibited rapid drift, as evidenced by the sharp drop in model performance shortly after deployment. By considering only the newly acquired data for each retraining, the 'New data only' strategy allowed the model to learn the latest patterns in the data and quickly adapt to changes. 

We observed that the 'New and some old data combined' strategy is more effective for chronological split (Strategies 7 and 9 in Table \ref{tab:Model Retraining Strategy Results}). This is because in this strategy, the training dataset of the current period includes the validation and test datasets from the previous period, ensuring that the model does not miss any data from a specific period, which is the case with the 'New data only' strategy. 

\subsection{Retraining Schedule}
We observed for all the strategies that the threshold based retraining worked better compared to the fixed interval retraining (Table \ref{tab:Model Retraining Strategy Results}). The reason for this is likely the data drift in the dataset, which affected the model's performance faster than the predetermined fixed interval. Future work could explore the optimal fixed interval depending on the characteristics of the dataset. 

\section*{Conclusions}
This work presents a framework of decision points for the development of a CT process. We mapped out strategies for including newly acquired data for model retraining and evaluation, and compared strategies for finetuning LMs in a CT environment. We show that different considerations during the implementation of a monitoring and  retraining strategy may have substantial impact on the overall model performance in a real-world application. The presented framework can be effectively adapted to different datasets and LM-based use cases beyond the ones presented here.

\section*{Limitations}

While our study investigated several retraining strategies for machine learning models, there are several limitations to our work. 

For the purposes of our study, we used a specific dataset with unique characteristics, and the optimal retraining strategy may differ for other datasets. However, the proposed methodology generalises well across datasets, and we recommend that such experiments should be performed before model deployment when possible to explore the effectiveness of different retraining strategies in different use cases.

The retraining strategies explored in the experiments were limited to a single architecture, and did not aim to include experiments where different model architectures are routinely compared during CT. While this is a valid avenue for future research, exploring multiple types of challengers in a CT environment might become rapidly impractical and costly.

Finally, an important assumption made in our study is that the metrics used for monitoring adequately reflect model performance in the MLC context. It should be noted that this assumption may not hold in all scenarios and should be revisited when applying monitoring strategies in different use cases. 

\section*{Ethics Statement}
Our research study has been guided by strong ethical values and procedures. Our objective is to evaluate the effectiveness of various retraining procedures and provide empirical evidence for others to learn from. In order to reduce potential sources of bias, we have taken care in designing our evaluation methodology. 

We recognize that working with language models is a computationally intensive process that can have a significant impact on the environment by leaving a large carbon footprint. Our research aims to help ML practitioners make informed decisions that can reduce the compute requirements and thus the carbon footprint. 

We are also aware that transparency is a crucial ethical factor. To ensure transparency, we have taken great care in accurately and clearly explaining our models, their performance, and the processes involved in our paper.

\bibliographystyle{unsrt}  
\bibliography{references,anthology}  

\begin{thebibliography}{10}

\bibitem{nigenda_amazon_2022}
David Nigenda, Zohar Karnin, Muhammad~Bilal Zafar, Raghu Ramesha, Alan Tan, Michele Donini, and Krishnaram Kenthapadi.
\newblock Amazon {SageMaker} model monitor: A system for real-time insights into deployed machine learning models.
\newblock In {\em Proceedings of the 28th {ACM} {SIGKDD} Conference on Knowledge Discovery and Data Mining}, {KDD} '22, pages 3671--3681. Association for Computing Machinery, 2022.

\bibitem{bayram_concept_2022}
Firas Bayram, Bestoun~S. Ahmed, and Andreas Kassler.
\newblock From concept drift to model degradation: An overview on performance-aware drift detectors.
\newblock {\em Knowledge-Based Systems}, 245:108632, 2022.

\bibitem{ackerman_automatically_2021}
Samuel Ackerman, Orna Raz, Marcel Zalmanovici, and Aviad Zlotnick.
\newblock Automatically detecting data drift in machine learning classifiers, 2021.

\bibitem{yu_automatic_2021}
Hang Yu, Tianyu Liu, Jie Lu, and Guangquan Zhang.
\newblock Automatic learning to detect concept drift, 2021.

\bibitem{katz2023natural}
Daniel~Martin Katz, Dirk Hartung, Lauritz Gerlach, Abhik Jana, and Michael~James Bommarito.
\newblock Natural language processing in the legal domain.
\newblock {\em Available at SSRN 4336224}, 2023.

\bibitem{rubin2012statistical}
Timothy~N Rubin, America Chambers, Padhraic Smyth, and Mark Steyvers.
\newblock Statistical topic models for multi-label document classification.
\newblock {\em Machine learning}, 88:157--208, 2012.

\bibitem{chalkidis-sogaard-2022-improved}
Ilias Chalkidis and Anders S{\o}gaard.
\newblock Improved multi-label classification under temporal concept drift: Rethinking group-robust algorithms in a label-wise setting.
\newblock In {\em Findings of the Association for Computational Linguistics: ACL 2022}, pages 2441--2454, Dublin, Ireland, May 2022. Association for Computational Linguistics.

\bibitem{gonzalez-carvajal_comparing_2021}
Santiago González-Carvajal and Eduardo~C. Garrido-Merchán.
\newblock Comparing {BERT} against traditional machine learning text classification, 2021.

\bibitem{chandola_outlier_2009}
Varun Chandola and Vipin Kumar.
\newblock Outlier detection : A survey.
\newblock {\em ACM Computing Surveys}, 41, 2009.

\bibitem{lu_learning_2019}
Jie Lu, Anjin Liu, Fan Dong, Feng Gu, João Gama, and Guangquan Zhang.
\newblock Learning under concept drift: A review.
\newblock {\em {IEEE} Transactions on Knowledge and Data Engineering}, 31(12):2346--2363, 2019.
\newblock Conference Name: {IEEE} Transactions on Knowledge and Data Engineering.

\bibitem{das_fairness_2021}
Sanjiv Das, Michele Donini, Jason Gelman, Kevin Haas, Mila Hardt, Jared Katzman, Krishnaram Kenthapadi, Pedro Larroy, Pinar Yilmaz, and Muhammad~Bilal Zafar.
\newblock Fairness measures for machine learning in finance.
\newblock {\em The Journal of Financial Data Science}, 3(4):33--64, 2021.

\bibitem{chakraborty_adversarial_2018}
Anirban Chakraborty, Manaar Alam, Vishal Dey, Anupam Chattopadhyay, and Debdeep Mukhopadhyay.
\newblock Adversarial attacks and defences: A survey, 2018.

\bibitem{raykar_data_2015}
Vikas~C. Raykar and Amrita Saha.
\newblock Data split strategiesfor evolving predictive models.
\newblock In Annalisa Appice, Pedro~Pereira Rodrigues, Vítor Santos~Costa, Carlos Soares, João Gama, and Alípio Jorge, editors, {\em Machine Learning and Knowledge Discovery in Databases}, Lecture Notes in Computer Science, pages 3--19. Springer International Publishing, 2015.

\bibitem{ben-zaken-etal-2022-bitfit}
Elad Ben~Zaken, Yoav Goldberg, and Shauli Ravfogel.
\newblock {B}it{F}it: Simple parameter-efficient fine-tuning for transformer-based masked language-models.
\newblock In {\em Proceedings of the 60th Annual Meeting of the Association for Computational Linguistics (Volume 2: Short Papers)}, pages 1--9, Dublin, Ireland, May 2022. Association for Computational Linguistics.

\bibitem{schulz_analysis_2019}
Claudia Schulz, Christian~M. Meyer, Jan Kiesewetter, Michael Sailer, Elisabeth Bauer, Martin~R. Fischer, Frank Fischer, and Iryna Gurevych.
\newblock Analysis of automatic annotation suggestions for hard discourse-level tasks in expert domains.
\newblock In {\em Proceedings of the 57th Annual Meeting of the Association for Computational Linguistics}, pages 2761--2772. Association for Computational Linguistics, 2019.

\bibitem{reimers_reporting_2017}
Nils Reimers and Iryna Gurevych.
\newblock Reporting score distributions makes a difference: Performance study of {LSTM}-networks for sequence tagging.
\newblock In {\em Proceedings of the 2017 Conference on Empirical Methods in Natural Language Processing}, pages 338--348. Association for Computational Linguistics, 2017.

\bibitem{liu2019roberta}
Yinhan Liu, Myle Ott, Naman Goyal, Jingfei Du, Mandar Joshi, Danqi Chen, Omer Levy, Mike Lewis, Luke Zettlemoyer, and Veselin Stoyanov.
\newblock Roberta: A robustly optimized bert pretraining approach.
\newblock {\em arXiv preprint arXiv:1907.11692}, 2019.

\bibitem{2017arXiv170201460S}
P.~{Szyma{\'n}ski} and T.~{Kajdanowicz}.
\newblock {A scikit-based Python environment for performing multi-label classification}.
\newblock {\em ArXiv e-prints}, February 2017.

\bibitem{klabjan_neural_2020}
Diego Klabjan and Xiaofeng Zhu.
\newblock Neural network retraining for model serving.
\newblock {\em arXiv preprint arXiv:2004.14203}, 2020.
\newblock Publication Title: {arXiv} e-prints {ADS} Bibcode: 2020arXiv200414203K Type: article.

\end{thebibliography}

\appendix
\counterwithin{figure}{section}

\section{Model Details and Hyperparameters}
\label{sec:appendix}
In this appendix, we provide details about the RoBERTa model architecture and the hyperparameters used in our experiments.

\subsection{RoBERTa Model Architecture}

RoBERTa (Robustly Optimized BERT Approach) is a pre-trained model on English language using a masked language modeling (MLM) objective. The architecture of RoBERTa is similar to that of BERT, with a few key differences in the pre-training process. RoBERTa uses larger batch sizes, longer training sequences, and dynamically changing mask patterns to improve the model's robustness and generalization. We have considered \href{https://huggingface.co/roberta-base}{roberta-base} model which has 12 transformer layers, with a hidden size of 768, and 12 attention heads. The model takes as input a sequence of tokens, with a maximum sequence length of 512, and outputs the one or more labels for the input.

\subsection{Hyperparameters}

We used the following hyperparameters for our experiments:

Learning rate: We used a learning rate of 1e-4.

Learning rate scheduler: We used a cosine learning rate scheduler

Warmup steps: Warmup steps are set as 10\% of the training size 

Batch size: We used a batch size of 180.

Optimizer: We have used AdamW optimizer

Evaluation Steps: Evaluation steps are set such that there are 5 evaluation steps for an epoch.

Max input size : 512 tokens ( To address the 512-token limit, we initially processed the documents using the TextRank algorithm (implemented in \href{https://radimrehurek.com/gensim_3.8.3/summarization/summariser.html}{Gensim}) to generate summaries of the text.)

Stopping criteria: We finetuned the RoBERTa model with early stopping criteria of 5 evaluation steps with an improvement threshold of 0.001 weighted F1-score and maximum of 35 epochs. 

Evaluation metric: Weighted-F1 Score

We acknowledge that there may be other combinations of hyperparameters that could achieve better results, and we encourage future research to explore this space.

\subsection{Hardware}

We used AWS g5.xlarge machine for all our experiments

The details of RoBERTa model architecture and hyperparameters used in our experiments are included to facilitate the transparency and reproducibility of our experiments and to enable future research in this area.

\section{Data Split Strategies}
\centering
\begin{figure}[H]
    \centering
    \includegraphics[width=12cm]{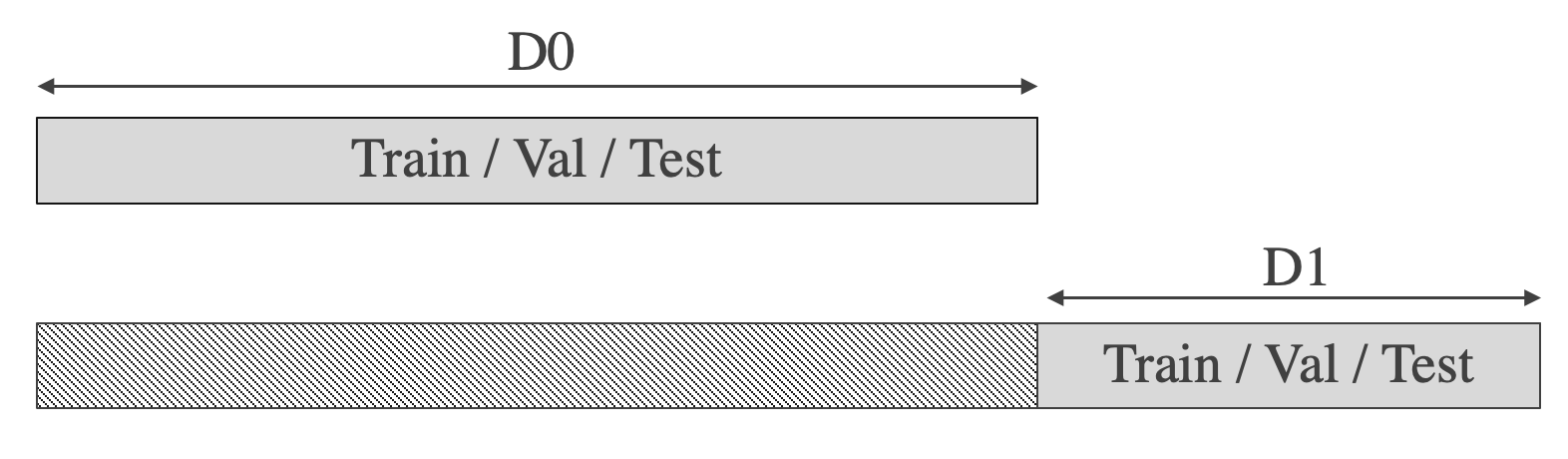}
    \caption{Stratified split with New data only}
    \label{fig:Stratified_split_with_rapid_data_refresh}
\end{figure}
\begin{figure}[H]
    \centering
    \includegraphics[width=12cm]{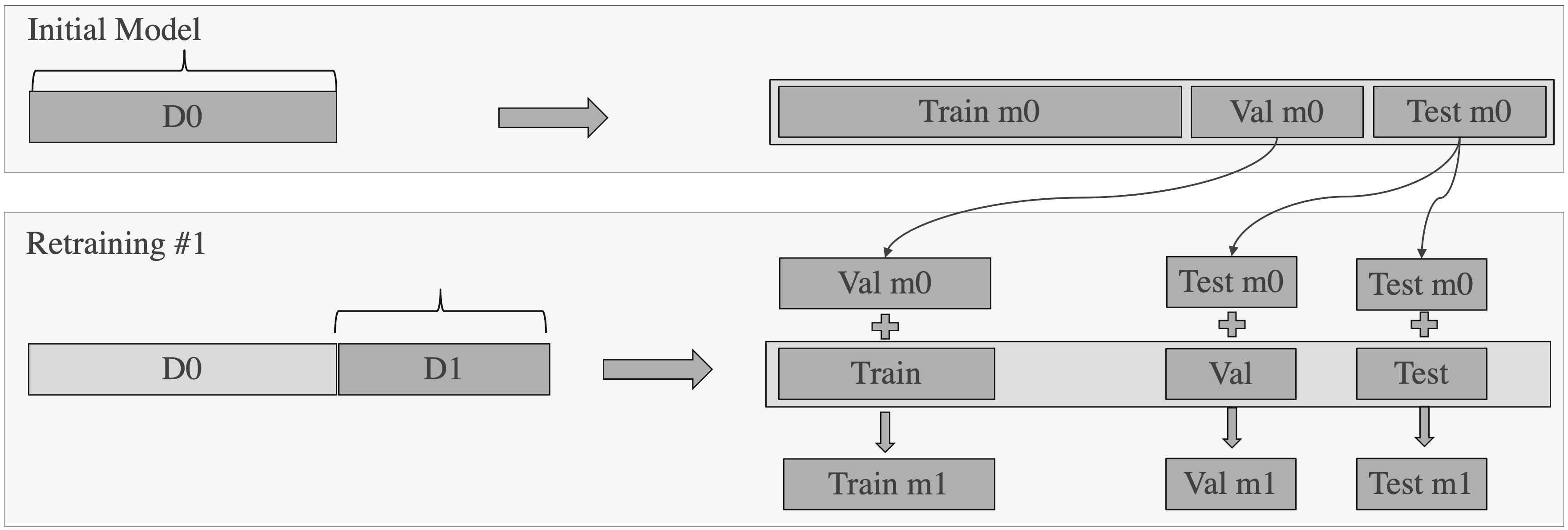}
    \caption{Stratified split with New and some old data combined}
    \label{fig:Stratified_split_with_waterfall_movement}
\end{figure}
\begin{figure}[H]
    \centering
    \includegraphics[width=12cm]{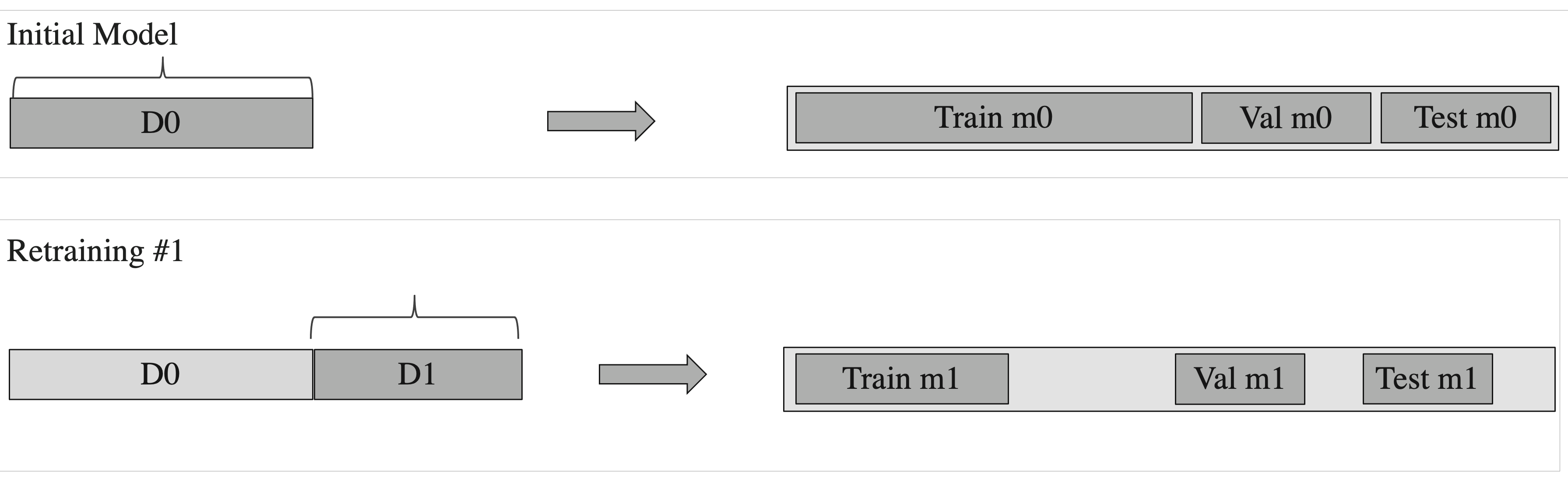}
    \caption{Chronological split with New data only}
    \label{fig:Chronological_split_with_rapid_data_refresh}
\end{figure}
\begin{figure}[H]
    \centering
    \includegraphics[width=12cm]{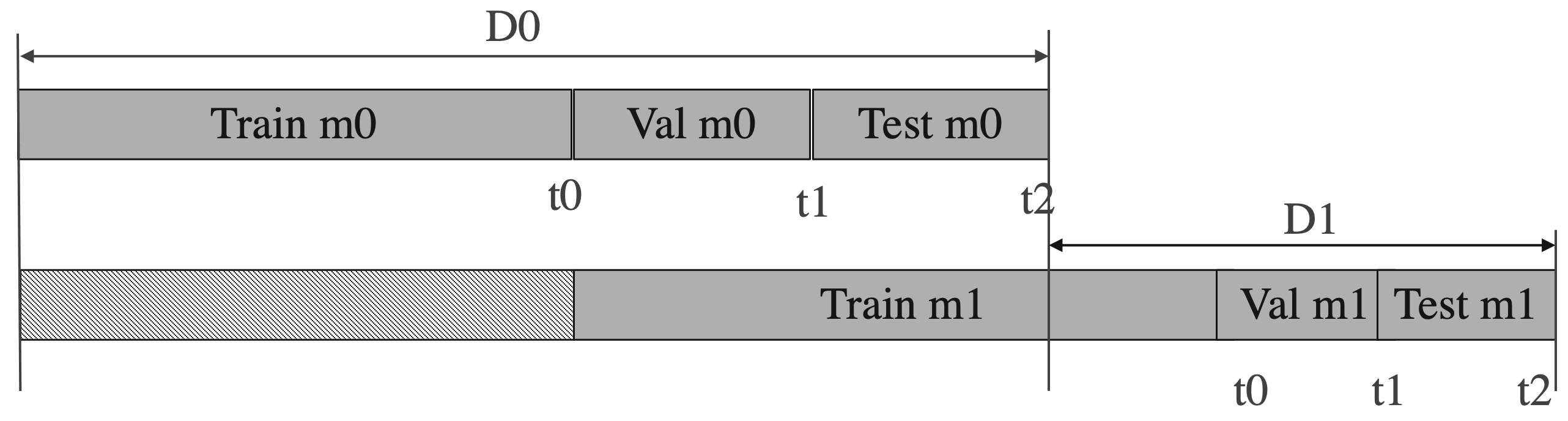}
    \caption{Chronological split with New and some old data combined}
    \label{fig:Chronological_split_with_moving_window}
\end{figure}

\end{document}